\documentclass[11pt]{article}
\usepackage{colortbl,booktabs}
\usepackage{ragged2e}
\usepackage{makecell}
\usepackage{tabularx}
\usepackage{textcomp}
\usepackage{tabu}
\usepackage{cuted}
\usepackage[shortlabels]{enumitem}
\usepackage{multirow}
\usepackage{setspace}
\usepackage{amsmath}
\usepackage[utf8]{inputenc}
\usepackage{amssymb}

\usepackage{arydshln}

\usepackage{algorithm}
\usepackage{algpseudocode}
\usepackage[preprint]{acl}

\usepackage{times}
\usepackage{latexsym}

\usepackage[T1]{fontenc}

\usepackage[utf8]{inputenc}

\usepackage{microtype}

\usepackage{inconsolata}

\usepackage{graphicx}

\usepackage[most]{tcolorbox}
\usepackage{scalerel}
\definecolor{myblue}{HTML}{4b8dbc}
\definecolor{mycolor}{HTML}{a7caea} 
\newtcolorbox{tocbox}{
  enhanced,
  breakable,
  colback=white,
  colframe=black,
  boxrule=0.4pt, 
  left=8pt,right=8pt,top=8pt,bottom=8pt,
}

\title{Focus-dLLM: Accelerating Long-Context Diffusion LLM Inference via Confidence-Guided Context Focusing}

\author{
  Lingkun Long$^{1}$, 
  Yushi Huang$^{2,3}$, 
  Shihao Bai$^{3}$, 
  Ruihao Gong$^{1,3}$, \\
  {\bf Jun Zhang$^{2}$, 
  Ao Zhou$^{1}$, 
  Jianlei Yang$^{1}$} \\
  $^1$Beihang University \quad $^2$Hong Kong University of Science and Technology \quad \\ $^3$SenseTime Research
}

\begin{document}
\maketitle
\begin{abstract}
Diffusion Large Language Models (dLLMs) deliver strong long-context processing capability in a non-autoregressive decoding paradigm. However, the considerable computational cost of bidirectional full attention limits the inference efficiency. Although sparse attention is promising, existing methods remain ineffective. This stems from the need to estimate attention importance for tokens yet to be decoded, while the unmasked token positions are unknown during diffusion. In this paper, we present \textbf{Focus-dLLM}, a novel training-free attention sparsification framework tailored
for accurate and efficient long-context dLLM inference. Based on the finding that token confidence strongly correlates across adjacent steps, we first design a \textit{past confidence-guided indicator} to predict unmasked regions. Built upon this, we propose a \textit{sink-aware pruning strategy} to accurately estimate and remove redundant attention computation, while preserving highly influential attention sinks. To further reduce overhead, this strategy reuses identified sink locations across layers, leveraging the observed cross-layer consistency. Experimental results show that our method offers more than $29\times$ lossless speedup under $32K$ context length. The code is publicly available at: \url{https://github.com/Longxmas/Focus-dLLM}.
\end{abstract}



\section{Introduction}\label{sec:intro}

Diffusion large language models (dLLMs)~\cite{bie2025llada20scalingdiffusionlanguage, gong2025scalingdiffusionlanguagemodels, arriola2025blockdiffusioninterpolatingautoregressive} have recently emerged as a compelling non-autoregressive paradigm for text generation, replacing left-to-right token emission with iterative denoising over a fixed-length sequence~\cite{li2022diffusion,gong2022diffuseq,austin2021structured,lou2023discrete,he2023diffusionbert}.
By updating multiple positions in parallel and leveraging bidirectional attention, dLLMs offer an appealing path toward higher decoding throughput while retaining strong generation quality.
Moreover, recent studies have substantially extended the context length of dLLMs~\cite{liu2025longlladaunlockinglongcontext,he2025ultralladascalingcontextlength}, demonstrating effective long-context extrapolation and scaling to long inputs.

Nevertheless, efficient long-context inference remains a key obstacle for the dLLM due to its \textit{non-autoregressive} decoding and bidirectional \textit{full} attention nature. Prior methods~\cite{wu2025fastdllmtrainingfreeaccelerationdiffusion,liu2025dllmcacheacceleratingdiffusionlarge,ma2025dkv} to address this challenge fall into two categories: (\textit{i}) \emph{Approximated} KV cache and (\textit{ii}) sparse attention. The former selectively refreshes KV states by exploiting strong redundancy between adjacent steps.  However, attention computation is still costly over the \textit{full} cached context. On the other hand, sparse attention~\cite{tang2024quest,xiao2024infllm, xu2025xattention,yuan2025nativesparseattentionhardwarealigned} offers a practical solution, but it often requires token importance estimation using the \emph{currently decoded} token as a query~\cite{zhang2023h2o,xiao2024infllm}. Since the positions to be decoded (unmasked) are not known in advance for dLLMs, recent works~\cite{song2025sparsedllmacceleratingdiffusionllms, huang2025masktokensprophetfinegrained} leverage inaccurate coarse-grained estimation, leading to suboptimal performance and limited efficiency. This paper, therefore, asks: \textit{Can we accurately predict the positions of the unmasked tokens and only retain necessary computation to achieve more effective long-context inference acceleration for dLLMs?}

To tackle this challenge, we first make an in-depth analysis to investigate the predictability of the unmasked tokens. In particular, we discover that the confidence scores at the same positions in two consecutive steps exhibit a strong positive correlation, and the positions of currently unknown tokens largely overlap with those that had the highest-confidence tokens in the previous step. Thus, unmasked positions for the current steps can be inferred from previous-step confidence. Besides, we also analyze the redundancy of attention patterns and observe that attention sink~\cite{xiao2023efficient,ruscio2025sinkinggeometricapproachattention}, which contributes significantly to the attention score in LLMs~\cite{bai2023qwentechnicalreport,touvron2023llamaopenefficientfoundation}, displays notable cross-layer consistency for dLLMs. This phenomenon suggests sink tokens can be identified at an intermediate depth. Therefore, we can directly reuse them without re-identification in deeper layers.

Motivated by the above findings, we propose Focus-dLLM, a training-free sparse attention framework with approximated KV cache, to accelerate long-context dLLM inference. To begin with, we introduce a \textit{past confidence-guided indicator} that uses confidence scores from step $t\!-\!1$ to predict the unmasked positions at step $t$, and then window-expands them to preserve semantic coherence. Next, we design a \textit{sink-aware pruning strategy} for diffusion decoding: Using the tokens within the positions predicted before as queries, we select only the most relevant tokens for attention while retaining step-wise attention sinks. Moreover, this approach shares the identified sink tokens across layers to further reduce additional overhead. Leveraging these novel techniques, our framework computes attention over the predicted unmasked queries and the selected necessary key-value pairs. As a result, it achieves considerable inference speedups without compromising performance throughout the dynamic decoding process. 

Our contributions are summarized as follows:
\begin{itemize}[leftmargin=*,nosep]
    \item We analyze diffusion inference dynamics and reveal a strong positive correlation of token confidence across adjacent denoising steps, together with dynamic and structured attention patterns in dLLMs.
    \item We propose Focus-dLLM, a novel training-free acceleration framework that consists of a past confidence-guided indicator for predicting the next unmasked positions with a sink-aware dynamic token pruning strategy for efficient sparse attention.
    \item Experiments show that Focus-dLLM achieves substantial speedups over baselines while preserving accuracy. For instance, it attains better-than-vanilla performance and delivers $2.05\times$ speedup over Fast-dLLM for UltraLLaDA at $32K$ context length.
\end{itemize}

\section{Related Work}
\noindent\textbf{Diffusion large language models.}
Diffusion large language models (dLLMs)~\cite{li2025survey, you2025llada, chen2025masked} have emerged as a promising non-autoregressive paradigm that enables parallel token generation via iterative denoising. Prior works explore both continuous-space diffusion for text~\cite{li2022diffusion,gong2022diffuseq} and discrete-token diffusion formulations~\cite{austin2021structured,lou2023discrete,he2023diffusionbert}. Recent masked diffusion LMs~\cite{nie2025llada, zhu2025llada15variancereducedpreference, ye2025dream7bdiffusionlarge} have been successfully scaled up, demonstrating competitive performance against autoregressive counterparts at billion-parameter scales. Besides, long-context capability~\cite{liu2025longlladaunlockinglongcontext,he2025ultralladascalingcontextlength} for dLLMs has also been explored, which pushes the context window up to ${\geq}16K$ tokens. 


\noindent\textbf{KV cache for dLLMs.}
Due to bidirectional attention and token states evolving across denoising steps, dLLMs cannot directly reuse standard KV cache, motivating a line of caching-based accelerations~\cite{ma2025dkv}. Fast-dLLM~\cite{wu2025fastdllmtrainingfreeaccelerationdiffusion} enables approximate KV reuse with block-wise strategies, while others~\cite{ma2025dkv, liu2025dllmcacheacceleratingdiffusionlarge,huang2025masktokensprophetfinegrained} exploit dLLM-specific redundancy to reduce repeated computation. More adaptive schemes ~\cite{jiang2025d2cacheacceleratingdiffusionbasedllms, nguyentri2025attentionneedkvcache} further refine cache update granularity and timing. Nevertheless, accurately identifying which tokens require refresh in the next step remains challenging, and long-context inference still incurs substantial computation overhead under caching mechanisms.

\noindent\textbf{Sparse attention for dLLMs}.
Attention sparsification~\cite{zhang2025sageattention, zhang2024sageattention2, zhang2025spargeattn}, orthogonal to the KV cache mechanism,  has also been explored to accelerate dLLM inference. Sparse-dLLM~\cite{song2025sparsedllmacceleratingdiffusionllms} proposes dynamic cache eviction for diffusion decoding, but it adopts coarse and suboptimal block-level metrics. SparseD~\cite{wang2025sparsed} reuses prior sparse patterns, yet it still relies on dense attention in early steps, restricting speedups. Moreover, these approaches do not account for the dynamic attention-sink behavior~\cite{xiao2023efficient} observed in dLLMs~\cite{rulli2025attentionsinksdiffusionlanguage}. In contrast, our dynamic KV cache compression scheme adapts to step-varying contextual needs while preserving attention sinks for more efficient and accurate long-context inference.

\section{Preliminaries}
\label{sec:preliminaries}

\noindent\textbf{Diffusion LLM inference.}\label{sec:dli}
Unlike autoregressive models that generate tokens sequentially, dLLMs generate text by iteratively denoising a fixed-length sequence.
Let $\mathcal{V}$ denote the vocabulary and $\texttt{[MASK]}\in\mathcal{V}$ the special mask token.
Given a prompt $\mathbf{p}=[p_1,\dots,p_M]$, inference initializes at step $0$ a length-$L$ sequence by appending $N=L-M$ masks:
\begin{equation}
\begin{array}{ll}
     \mathbf{x}^{(T)}=[\underbrace{p_1,\dots,p_M}_{\text{Prompt}},\underbrace{\texttt{[MASK]},\dots,\texttt{[MASK]}}_{N=L-M}],
\end{array}
\end{equation}

Let $\mathcal{M}^{(t)}$ denote the set of masked positions at denoising step $t$, where $\mathcal{M}^{(0)}=\{M+1,\dots,L\}$ at initialization.
The decoding process then iterates from $t=0$ to $T-1$. In step $t$, given the current sequence $\mathbf{x}^{(t)}$, the model $f_\theta$ produces a conditional distribution $p(x_i \mid \mathbf{x}^{(t)})$ for each masked position $i\in\mathcal{M}^{(t)}$.
Then, a confidence-driven strategy~\cite{nie2025llada,ye2025dream7bdiffusionlarge} computes the predicted token $\hat{x}_i^{(t)}$ and its corresponding confidence score $c_i^{(t)}$ for each masked position $i$:
\begin{equation}
\begin{aligned}
\hat{x}_i^{(t)} &= \arg\max_{v\in\mathcal{V}} p(x_i=v \mid \mathbf{x}^{(t)}), \\
c_i^{(t)} &= \max_{v\in\mathcal{V}} p(x_i=v \mid \mathbf{x}^{(t)}).
\end{aligned}
\label{eq:confidence}
\end{equation}
Last, this strategy unmasks the highest-confidence positions while remasking the rest.

\noindent\textbf{Approximate KV cache in dLLMs.} Bidirectional attention makes the KV cache mechanism not applicable for dLLMs. To reduce computation costs, recent studies~\cite{wu2025fastdllmtrainingfreeaccelerationdiffusion,liu2025dllmcacheacceleratingdiffusionlarge,ma2025dkv} exploit \emph{Approximate} KV cache, which updates KV states for a selected subset of tokens while reusing cached states for the rest. 
Formally, let $\mathcal{U}^{(t)}$ be the token indices refreshed at step $t$. The Key state $\mathbf{K}_i^{(t)}$ (and similarly $\mathbf{V}_i^{(t)}$) is
\begin{equation} \mathbf{K}_i^{(t)} = \begin{cases} f_{K}(\mathbf{x}^{(t)})_i, & i \in \mathcal{U}^{(t)} \quad (\text{Compute}) \\ \widetilde{\mathbf{K}}_i, & i \notin \mathcal{U}^{(t)} \quad (\text{Reuse}) \end{cases}, \label{eq:kv_cache} \end{equation}
where $\widetilde{\mathbf{K}}_i$ denotes the cached state from the previous iteration. $f_{K}(\mathbf{x}^{(t)})_i$ is the current computed Key state, which is also used to update the cache.

\section{Motivation}
In this section, we investigate the token-confidence consistency and attention patterns tailored for dLLMs. Both of them inspire the core design of our Focus-dLLM.

\subsection{Temporal Consistency of Confidence}\label{sec:tcc}
\begin{figure}[t]
  \includegraphics[width=\columnwidth]{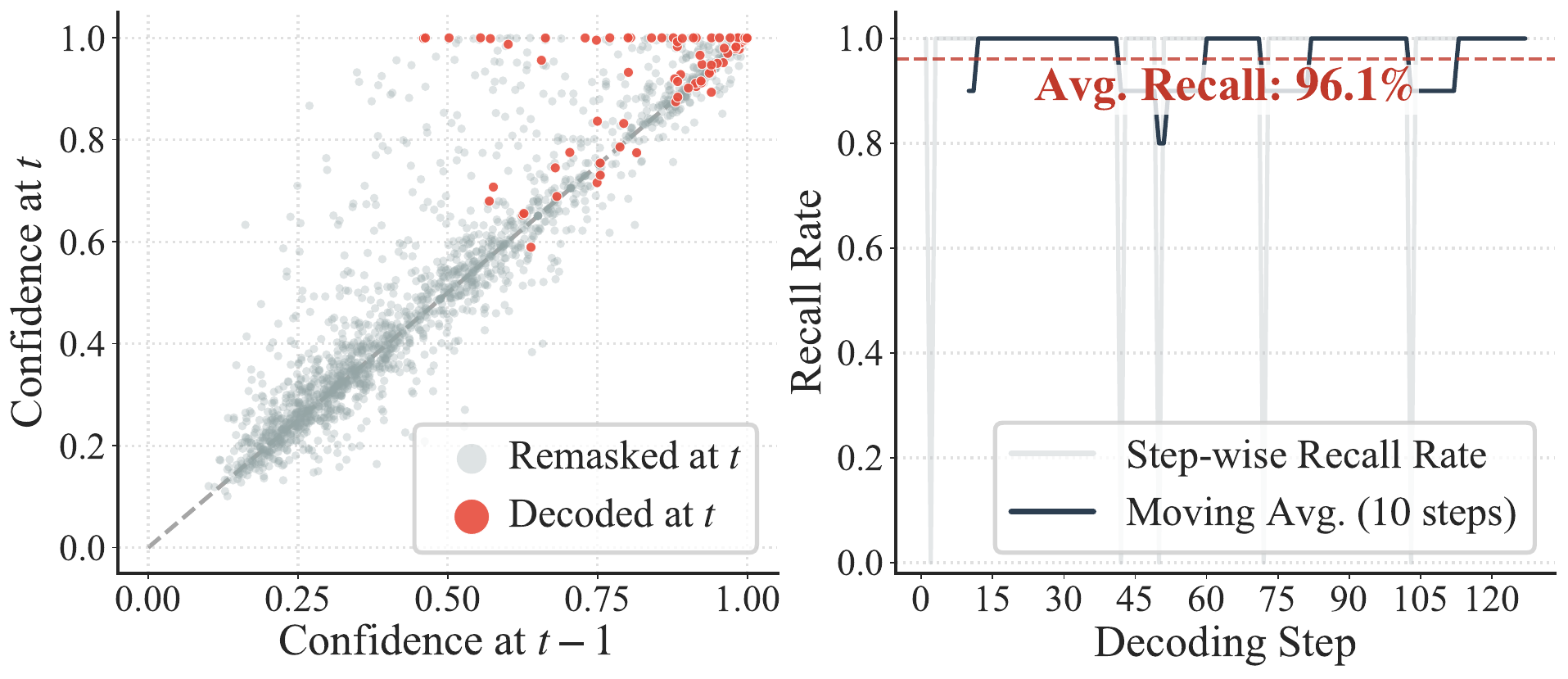 }
  \caption{Confidence dynamics analysis for LLaDA-8B-Instruct~\cite{nie2025llada} on GSM$8K$~\cite{cobbe2021gsm8k} ($L=76$, $N=128$, and $T=128$). (\textit{Left}) Confidence score correlation between adjacent steps. (\textit{Right}) Step-wise recall rates of predicting the unmasked tokens at $t$ using the remasked tokens with top-$4$ highest confidence scores at $t-1$.}
  \label{fig:confidence_guided}
  \vspace{-0.1in}
\end{figure}

For dLLMs, effectively assessing the redundancy of attention computation \emph{w.r.t.} tokens that are poised to be unmasked first requires locating these tokens in advance. To achieve this, we conduct a pivotal study related to their confidence score $c^{(t)}_i$. As illustrated in Figure~\ref{fig:confidence_guided} (\textit{Left}), $c^{(t)}_i$ and $c^{(t-1)}_i$ correlate strongly in a positive manner. Also, the tokens that are to be decoded (unmasked) at $t$ present a similarly high-confidence level in the preceding step $t-1$. To quantitatively explore this relationship, we select the top-$4$ remasked tokens (\emph{i.e.}, \texttt{[MASK]} at $t$) with the highest confidence scores at \(t-1\) and evaluate their overlap with the tokens decoded at the subsequent step \(t\). As a result, Figure~\ref{fig:confidence_guided} (\textit{Right}) shows a remarkably high average recall ($96.1\%$) across decoding steps. These observations support the following key claim:

\begin{tcolorbox}
The substantial overlap in confidence distributions reveals that tokens unmasked at $t$ can be reliably located according to the confidence of tokens at the prior step $t-1$.
\end{tcolorbox}

\subsection{Spatial Consistency of Attention Sinks}\label{sec:dsap}
\begin{figure}[t]
  \includegraphics[width=\columnwidth]{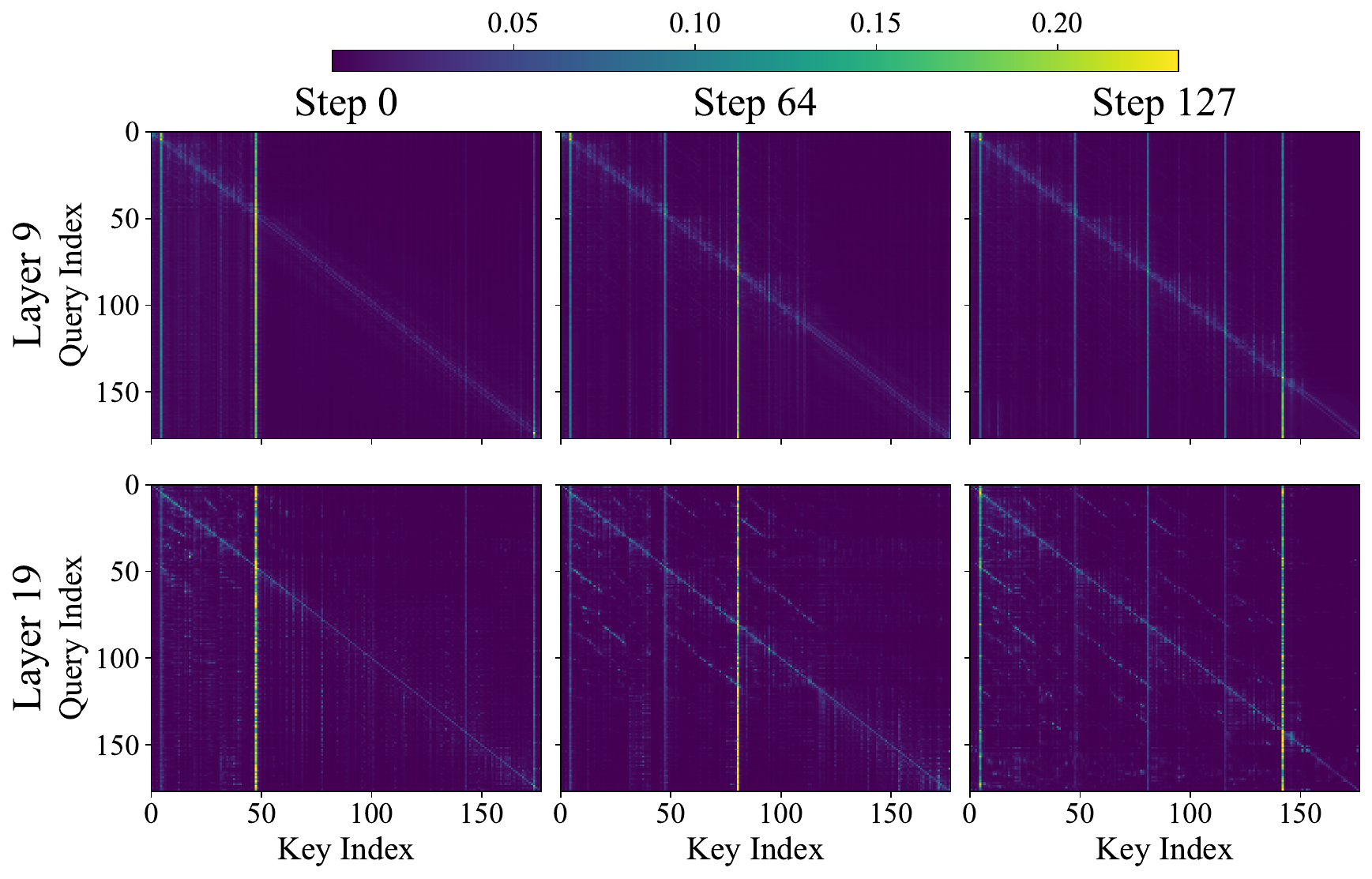 }
  \caption{Attention patterns across decoding steps and layers in LLaDA-8B-Instruct~\cite{nie2025llada} ($L=49$, $N=128$, $T=128$). More visual results can be found in the Appendix.}
  \label{fig:attn_map}
  \vspace{-0.2in}
\end{figure}

In this part, we explore the properties and variations of attention patterns for dLLMs. Similar to prior studies~\cite{song2025sparsedllmacceleratingdiffusionllms, rulli2025attentionsinksdiffusionlanguage}, as depicted in Figure~\ref{fig:attn_map}, we also find that:  (\emph{i}) Attention maps exhibit strong locality, concentrating near the diagonal and favoring nearby context. (\emph{ii}) Attention sinks (bright vertical bands), which strongly influence semantic continuity~\cite{xiao2023efficient, gu2025attentionsinkemergeslanguage}, emerge and evolve across denoising steps. Due to the dynamics of these sinks, it is necessary to repeatedly identify their location to preserve them in high-performing sparse attention. Despite this, we fortunately discovered a structured inter-layer consistency for attention sinks. To be specific, the index of attention sinks across different layers (\emph{e.g.}, Layer 9 \emph{vs}. Layer 19 in Figure~\ref{fig:attn_map}) typically matches. Therefore, we believe that:
\begin{tcolorbox}
Attention sinks with strong cross-layer consistency enable reliable identification at a certain intermediate depth, and the results can be reused for deeper layers to eliminate redundant computation.
\end{tcolorbox}

\section{Focus-dLLM}

\begin{figure*}[t]
\includegraphics[width=2.0\columnwidth]{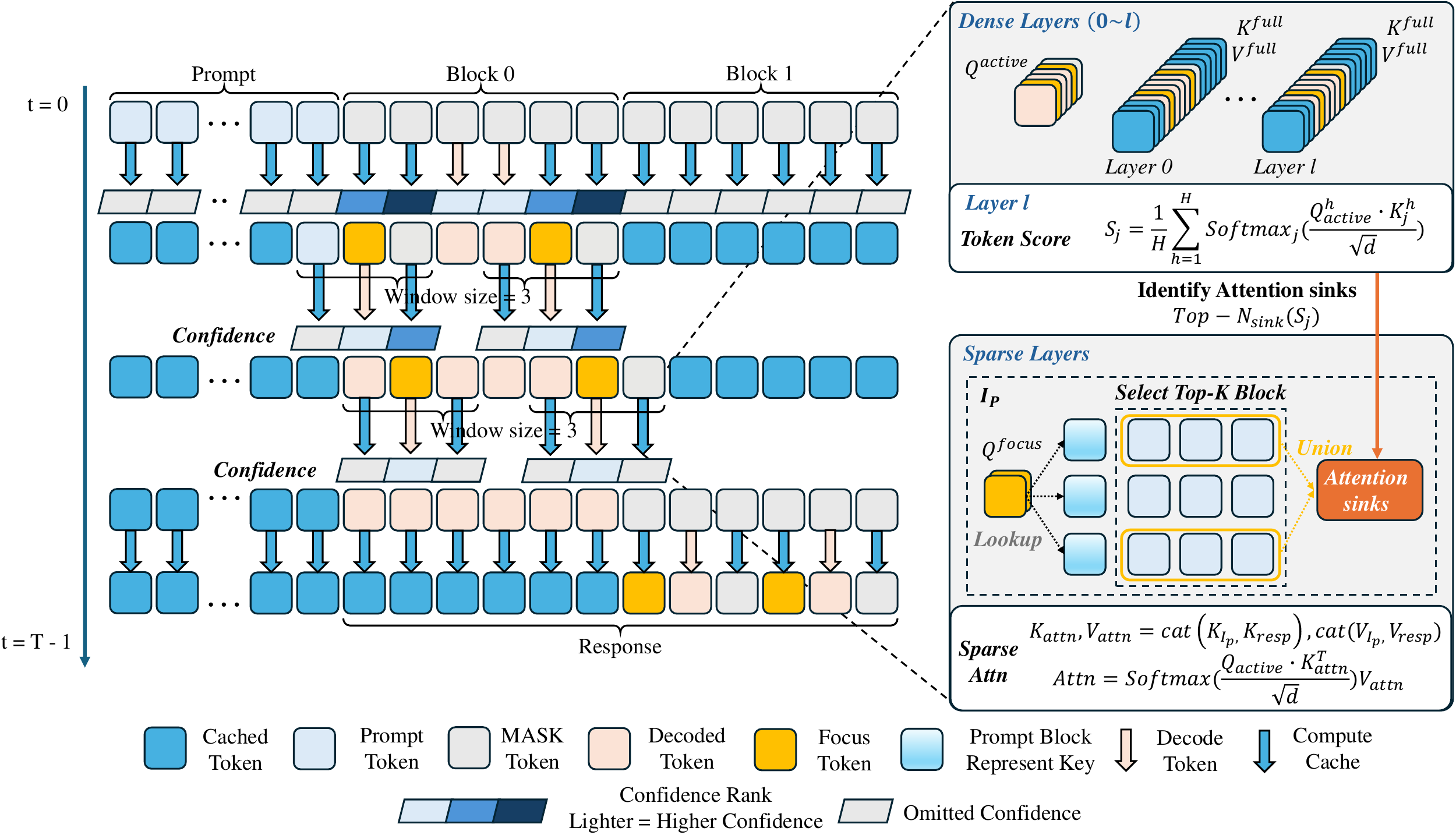 }
  \caption{Overview of Focus-dLLM. We predict unmasked positions at the current step using previous confidence scores. These positions act as queries to retrieve relevant prompt blocks, where attention is computed over the union of these blocks and dynamically identified attention sinks.}
  \label{fig:overview}
  \vspace{-0.1in}
\end{figure*}

\subsection{Framework Overview}

In this section, we present the inference workflow of Focus-dLLM (Figure~\ref{fig:overview}). Following Wu \textit{et al.}~\cite{wu2025fastdllmtrainingfreeaccelerationdiffusion}, we adopt the KV caching mechanism with the semi-autoregressive remasking strategy~\cite{nie2025llada} (\emph{i.e.}, non-autoregressive unmasking within each block and autoregressive block-wise inference from left block to right block) for dLLMs. 

For all blocks, Focus-dLLM performs a full cache refresh at each block entry step, which is commonly adopted in prior works~\cite{wu2025fastdllmtrainingfreeaccelerationdiffusion, song2025sparsedllmacceleratingdiffusionllms}. For the other denoising steps, we use our proposed sparse attention pipeline to systematically prune redundancy:
\begin{itemize}
    \item \textit{Section~\ref{sec:pcgi}:} Inspired by Section~\ref{sec:tcc}, we use confidence scores from the previous step to predict masked positions that are likely to be decoded, which provide focused queries for redundancy estimation in the latter Key/Value pruning. Additionally, guided by the locality pattern in Section~\ref{sec:dsap}, we expand these positions to local windows to form an active Query set and exclude the remaining Query tokens to compute the attention.
    \item \textit{Section~\ref{sec:sasa}:} We accelerate inference via a sink-aware sparse attention mechanism. Since shallow layers are more sensitive to sparsification~\cite{huang2025masktokensprophetfinegrained}, we treat the initial layers as \emph{dense layers} with full attention. For subsequent \emph{sparse layers}, we reuse the locations of attention sinks identified at the last dense layer. Finally, we apply dynamic block-wise pruning to Key/Value states of the prompt to keep the most relevant history, while retaining recognized sinks and all response tokens to preserve semantic coherence.
\end{itemize}

\subsection{Past Confidence-Guided Indicator}\label{sec:pcgi}

Motivated by the temporal consistency analysis in Section~\ref{sec:tcc}, we introduce a \textit{past confidence-guided indicator}, which adopts the confidence derived from step $t-1$ to accurately inform the tokens that are likely to be unmasked at step $t$. To be specific, among all positions $\mathcal{M}^{(t)}$ that remain in the \texttt{[MASK]} state within the current decoding block at $t$, we rank them by their prior confidence scores $c_j^{(t-1)}$ and select top-$k$ indices as the candidate set $\mathcal{I}_{\text{focus}}$ to predict the future unmasked positions at $t$:
\begin{equation}
    \mathcal{I}_{\text{focus}}= \left\{i\mid c_i^{(t-1)}\in \text{top-}k\big(\{c^{(t-1)}_j\}_{j\in \mathcal{M}^{(t)}}\big)\right\},
    \label{eq:candidate}
\end{equation}
where $k=\lfloor\rho n^{(t)}\rceil$. $n^{(t)}$ is the number of tokens to be unmasked and $\rho$ is a pre-defined prediction expansion factor. By leveraging the candidate set $\mathcal{I}_{\text{focus}}$, we can precisely determine the relevant history to prune redundant attention computation in the next subsection. 

In addition, as discussed in Section~\ref{sec:dsap}, the attention mechanism in dLLMs exhibits a clear locality property, meaning that token representations depend strongly on nearby semantic context, while distant tokens typically contribute little. To leverage this property for computation savings, we propose a window expansion strategy that disregards the distant tokens and only preserves local windows for currently decoded Query tokens (positions in $\mathcal{I}_{\text{focus}}$) for attention computation. The position set corresponding to the union of windows is given as:
\begin{equation}
    \mathcal{I}_{\text{active}}= \bigcup_{i \in \mathcal{I}_{\text{focus}}}\left\{l\mid i - \lfloor w/2 \rfloor \le l \le i + \lfloor w/2 \rfloor \right\},
    \label{eq:candidate}
\end{equation}
where $w$ is the window size.

\subsection{Sink-Aware Sparse Attention}\label{sec:sasa}

Performing attention over the entire long-context history remains the primary computational bottleneck for inference. To address this, we propose a sink-aware sparse attention strategy that selectively retains only the most critical history for diffusion decoding.

\noindent\textbf{Dynamic attention sinks identification.}
While retaining attention sinks is crucial for preserving generation quality~\cite{xiao2024efficientstreaminglanguagemodels, rulli2025attentionsinksdiffusionlanguage}, existing sparse approaches for dLLMs~\cite{song2025sparsedllmacceleratingdiffusionllms} typically overlook this, thereby risking the discard of tokens pivotal for generation quality~\cite{rulli2025attentionsinksdiffusionlanguage}. Motivated by our observation in Section~\ref{sec:dsap}, we propose to explicitly identify and retain them. Crucially, this strategy shares the identified sink tokens across layers, avoiding redundant re-calculation at every depth.

Specifically, we designate the first $l_{\text{dense}}$ layers as dense layers that perform full attention. Due to the cross-layer consistency observed in Section~\ref{sec:dsap}, we utilize the attention distribution at the cut-off layer $l_{\text{dense}}$ as a reliable probe to identify globally salient tokens for the subsequent sparse layers. Let $\mathcal{I}_{\text{active}}$ denote the active token set obtained from Section~\ref{sec:pcgi}.
We define the aggregated query representation over active tokens as $Q_{\mathcal{I}_{\text{active}}}$.
The importance score of each token $j$ is computed as:
\begin{equation}
S_j = \frac{1}{H} \sum_{h=1}^{H}
\operatorname{Softmax}_{j}
\left(
\frac{Q_{\mathcal{I}_{\text{active}}}^{h} \cdot K_{j}^{h}}{\sqrt{d}}
\right),
\end{equation}
where $H$ denotes the number of attention heads.

We then select the top-$N_{\text{sink}}$ tokens with the highest scores to form the dynamic attention sink set, denoted as $\mathcal{I}_{\text{sink}} = \operatorname{Top-}N_{\text{sink}}(S_j)$.

\begin{table*}[t]
\centering
\scriptsize
\setlength{\tabcolsep}{2pt}
\renewcommand{\arraystretch}{1.15}
\caption{Performance comparison on LongBench~\cite{bai2024longbench}. \textbf{Bold} indicates the best performance among acceleration methods, and \underline{underlined} indicates the second best.}
\label{tab:longbench_results}
 \resizebox{\linewidth}{!}{
\begin{tabu}{l*{14}{c}}
\toprule
\multirow{5}{*}{Method}
& \multicolumn{2}{c}{Single-Doc. QA}
& \multicolumn{3}{c}{Multi-Doc. QA}
& \multicolumn{2}{c}{Summarization}
& \multicolumn{2}{c}{Few-shot Learning}
& \multicolumn{2}{c}{Synthetic}
& \multicolumn{2}{c}{Code}
& \multirow{5}{*}{Ave.\ Score}
\\
\cmidrule(lr){2-3}\cmidrule(lr){4-6}\cmidrule(lr){7-8}\cmidrule(lr){9-10}\cmidrule(lr){11-12}\cmidrule(lr){13-14}
& \rotatebox{60}{Qasper}
& \rotatebox{60}{MF-en}
& \rotatebox{60}{HotpotQA}
& \rotatebox{60}{2WikiMQA}
& \rotatebox{60}{Musique}
& \rotatebox{60}{GovReport}
& \rotatebox{60}{QMSum}
& \rotatebox{60}{TREC}
& \rotatebox{60}{TriviaQA}
& \rotatebox{60}{Lsht}
& \rotatebox{60}{PRe}
& \rotatebox{60}{Lcc}
& \rotatebox{60}{RB-P}
& \\
\midrule

\multicolumn{15}{c}{\textit{UltraLLaDA}~\cite{he2025ultralladascalingcontextlength}}\\
\midrule

Vanilla
& 19.14 & 25.87
& 16.27 & 18.00 & 12.08
& 32.83 & 22.48 
& 80.00 & 91.58
& 41.00 & 96.75
& 68.23 & 59.50
& 44.90
\\
\midrule

Fast-dLLM
& \underline{18.34} & \textbf{29.90}
& 17.03 & 17.11 & 13.36
& \underline{30.05} & 22.89 
& \textbf{79.50} & 91.03
& \textbf{42.00} & 94.75
& \underline{67.50} & 58.10
& 44.74
\\

Sparse-dLLM
& 18.04 & 27.26
& \underline{20.59} & \underline{17.88} & \underline{13.67}
& 29.95 & \textbf{23.57} 
& 76.50 & \textbf{91.93}
& \underline{41.50} & \textbf{97.12}
& \underline{67.50} & 57.72
& \underline{44.86}
\\

SparseD
& \textbf{19.09} & 25.87
& 15.45 & 18.04 & 11.92
& \textbf{32.64} & \underline{22.50} 
& \textbf{79.50} & \underline{90.70}
& \underline{41.50} & \underline{96.79}
& \textbf{68.10} & \textbf{59.02}
& 44.70
\\

\rowcolor{mycolor!30}
Focus-dLLM
& 17.02 & \underline{29.11}
& \textbf{22.47} & \textbf{21.49} & \textbf{20.20}
& 26.75 & 21.45
& \underline{77.00} & 90.78
& 41.00 & 95.73
& 66.72 & 57.14
& \textbf{45.14}
\\

\midrule
\multicolumn{15}{c}{\textit{Dream-7B-Instruct}~\cite{ye2025dream7bdiffusionlarge}}\\

\midrule
Vanilla
& 35.58 & 40.49
& 41.59 & 42.10 & 23.36
& 23.51 & 19.66
& 71.50 & 87.34
& 15.75 & 32.50
& 63.79 & 62.23
& 43.03
\\
\midrule

Fast-dLLM
& 37.54 & \textbf{43.24}
& 35.56 & 35.74 & 17.97
& \underline{21.14} & 19.57 
& \underline{70.50} & 88.25
& 16.75 & \underline{46.17}
& 62.21 & 61.11
& 42.75
\\

Sparse-dLLM
& 37.50 & \underline{43.23}
& 36.83 & 34.97 & 17.05
& 20.60 & \textbf{20.05} 
& 70.00 & \underline{88.38}
& \underline{17.00} & \textbf{46.50}
& \underline{62.80} & \underline{61.20}
& 42.78
\\

SparseD
& \textbf{37.66} & 40.96
& \textbf{41.39} & \textbf{41.61} & \textbf{24.17}
& \textbf{23.51} & \underline{19.66}
& \textbf{72.50} & 86.85
& 15.75 & 36.83
& \textbf{63.86} & \textbf{61.98} & \textbf{43.59}
\\

\rowcolor{mycolor!30}
Focus-dLLM
& \underline{37.38} & 41.96
& \underline{38.96} & \underline{38.56} & \underline{18.05}
& 21.06 & 19.26
& 70.00 & \textbf{88.76}
& \textbf{17.25} & 44.25
& 60.62 & 60.50
& \underline{42.82}
\\

\bottomrule
\end{tabu}}
\end{table*}

\noindent\textbf{Block-wise token pruning.}
To accelerate inference while maximizing GPU efficiency, we implement block-wise token pruning to reduce computational overhead. Specifically, we partition the prompt tokens into contiguous blocks and assign each block a lightweight representative key, computed as the mean of the Key states within the block,
$\bar{K}_b = \text{Mean}_{j \in \text{Block}_b}(K_j).$

At timestep $t$, we estimate the relevance between the predicted candidate queries and each prompt block by aggregating their attention interactions. Concretely, for each block $b$, we compute a relevance score as
\begin{equation}
R_b = \frac{1}{H} \sum_{h=1}^{H}
\left(
Q_{\mathcal{I}_{\text{focus}}}^{h} \cdot \bar{K}_{b}^{h}
\right),
\end{equation}
where $\mathcal{I}_{\text{focus}}$ denotes the predicted candidate set obtained in Section~\ref{sec:pcgi}.

Based on these relevance scores, we select the top
$
C = \lfloor \alpha \cdot N_{\text{total\_blocks}} \rfloor
$
blocks to form the set of relevant blocks,
$
\mathcal{B}_{\text{relevant}} = \operatorname{Top}\text{-}C(R_b).
$
The final attention index set is constructed as the union of dynamically identified attention sinks and tokens within the selected relevant prompt blocks:
\begin{equation}
\mathcal{I}_{p}
=
\mathcal{I}_{\text{sink}}
\cup
\bigcup_{b \in \mathcal{B}_{\text{relevant}}}
\{\, i \mid i \in \text{Block}_b \,\}.
\end{equation}

Using this index set, we perform sparse attention by gathering keys and values exclusively from the selected prompt tokens and the  response tokens. Specifically, for the active queries $\mathcal{Q}_{\mathcal{I}_{\text{active}}}$, the effective Key-Value pairs are formed as:
$
K_{\text{attn}} = \mathrm{concat}(K_{\mathcal{I}_{p}}, K_{\text{resp}}),
\qquad
V_{\text{attn}} = \mathrm{concat}(V_{\mathcal{I}_{p}}, V_{\text{resp}}).
$
The resulting sparse attention is then computed as:
\begin{equation}
\mathrm{Attn}
=
\mathrm{Softmax}\!\left(
\frac{
Q_{\mathcal{I}_{\text{active}}}
K_{\text{attn}}^{\top}
}{\sqrt{d}}
\right)
V_{\text{attn}}.
\end{equation}

\section{Experiments}

\subsection{Experiments Settings}
\label{sec:exp_settings}

\noindent\textbf{Models.} We evaluate our method on two representative diffusion LLMs: UltraLLaDA~\cite{he2025ultralladascalingcontextlength} and Dream-7B-Instruct~\cite{ye2025dream7bdiffusionlarge}. 

\noindent\textbf{Baselines.} We compare Focus-dLLM against standard native inference (Vanilla) and three dLLM acceleration frameworks: Fast-dLLM~\cite{wu2025fastdllmtrainingfreeaccelerationdiffusion}, SparseD~\cite{wang2025sparsed}, and Sparse-dLLM~\cite{song2025sparsedllmacceleratingdiffusionllms}. 

\noindent\textbf{Benchmarks.} To comprehensively assess long-context capabilities, we conduct evaluations on LongBench~\cite{bai2024longbench}, a widely adopted benchmark specifically designed for multi-task long-context understanding. 

\noindent\textbf{Implementation details.} 
All experiments were conducted on NVIDIA H200 GPUs using OpenCompass~\cite{2023opencompass}. To ensure a fair comparison, all baselines utilize the recommended configurations provided in their official implementations. Specifically, for SparseD, we set $skip=20\%$, $ratio=30\%$, and $block\_size=128$; for Sparse-dLLM, we use retention ratio $r=0.5$ and kernel size $s=3$. For the Focus-dLLM setup, we adopt identical hyperparameters for both UltraLLaDA and Dream: we set prediction expansion factor $\rho=4$, window size $w=8$, dense layers $l_{\text{dense}} = 6$, and sparsity ratio $\alpha = 0.5$. Additionally, the number of sink tokens is set to $N_{\text{sinks}} = 0.01 \times M$, where $M$ denotes the prompt length, and prompt block size = 64. Additional details for datasets, models, and methods are provided in the Appendix~\ref{sec:impl_details}.

\subsection{Main Results}

\begin{figure*}[!ht]
\includegraphics[width=\linewidth]{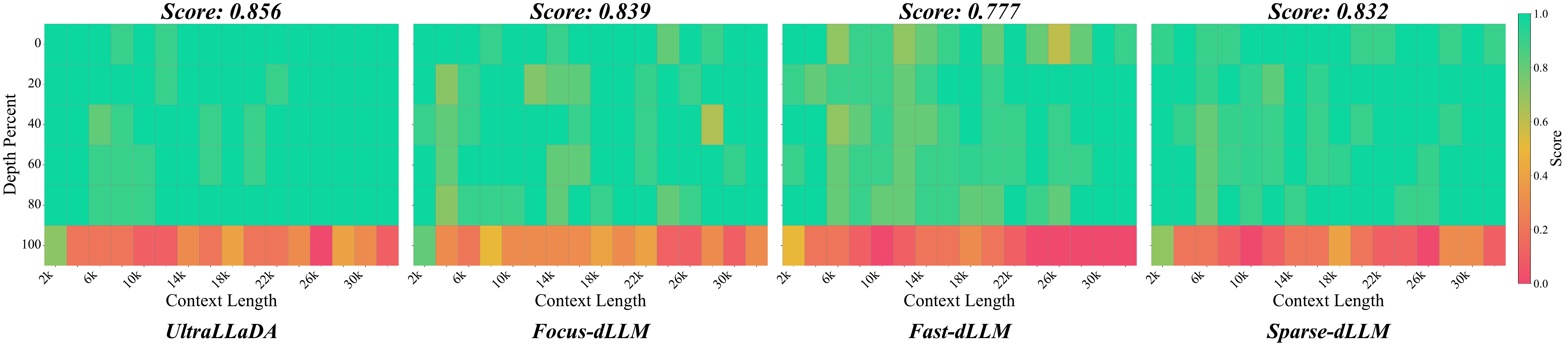} 
    \caption{Niah~\cite{kamradt2023niah} results on UltraLLaDA~\cite{he2025ultralladascalingcontextlength} under long-context settings with a maximum context length of $32K$ across different layer depths.}
\label{fig:niah}
\end{figure*}

\begin{figure*}[!ht]    
\includegraphics[width=0.95\linewidth]{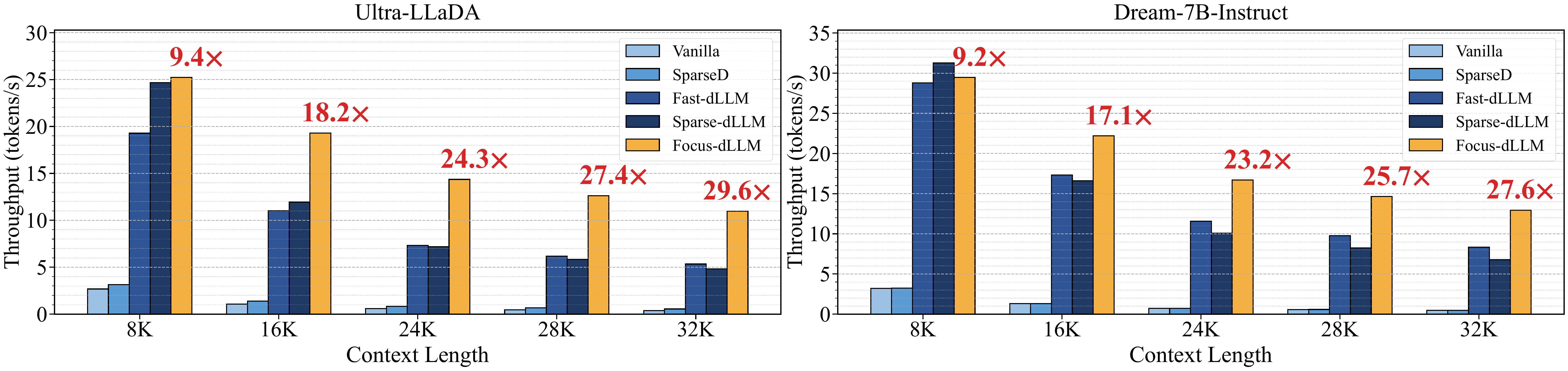} 
\centering
\caption{Efficiency evaluation. Comparison of decoding throughput (tokens/s) on UltraLLaDA~\cite{he2025ultralladascalingcontextlength} (\textit{Left}) and Dream-7B-Instruct~\cite{ye2025dream7bdiffusionlarge} (\textit{Right}) across varying context lengths. Red numbers indicate the speedup ratio of Focus-dLLM relative to the Vanilla baseline.}
    \label{fig:efficiency}
\end{figure*}

\noindent{\textbf{Accuracy.} 
As presented in Table~\ref{tab:longbench_results}, Focus-dLLM demonstrates robust performance across both evaluated diffusion models. 
On UltraLLaDA~\cite{he2025ultralladascalingcontextlength}, our method achieves the highest average score, outperforming the Vanilla baseline and all competing acceleration frameworks.
On Dream-7B-Instruct~\cite{ye2025dream7bdiffusionlarge}, Focus-dLLM again surpasses Spare-dLLM~\cite{song2025sparsedllmacceleratingdiffusionllms} and Fast-dLLM~\cite{wu2025fastdllmtrainingfreeaccelerationdiffusion}, performing on par with the Vanilla baseline. While its accuracy is marginally lower than SparseD~\cite{wang2025sparsed}, Focus-dLLM offers a compelling advantage in efficiency, achieving up to a $19.95\times$ speedup at a $32K$ (with 1K denoting 1024 tokens) context length (as shown in Figure~\ref{fig:efficiency}). This highlights our method's superior balance between performance and inference speed, establishing it as a more practical solution.

\noindent\textbf{Niah experiments.}
Figure~\ref{fig:niah} reports Niah~\cite{kamradt2023niah} results on
UltraLLaDA~\cite{he2025ultralladascalingcontextlength} under long-context settings
with a maximum context length of $32K$. Focus-dLLM achieves overall higher scores
than Fast-dLLM~\cite{wu2025fastdllmtrainingfreeaccelerationdiffusion} and
Sparse-dLLM~\cite{song2025sparsedllmacceleratingdiffusionllms} across layers,
and attains better accuracy than the vanilla baseline at the deepest layer,
demonstrating strong needle-in-a-haystack retrieval.

\noindent\textbf{Efficiency.}
We evaluate the scalability of Focus-dLLM by measuring throughput across context lengths from $8K$ to $32K$ context length, both the generation length and generation steps are fixed at 256. As shown in Figure~\ref{fig:efficiency}, our method consistently outperforms all baselines, with the speedup ratio over Vanilla notably expanding as context grows—from $9.4\times$ at $8K$ context length to $29.6\times$ at $32K$ context length. This trend can be attributed to the reduction of redundant attention computation, which tends to incur more significant overhead as sequences lengthen. Consequently, Focus-dLLM maintains superior efficiency and surpasses existing frameworks like Fast-dLLM by up to $2.05\times$ at $32K$ context length.

\noindent\textbf{Accuracy \emph{vs}. efficiency.}
\begin{figure}[t]    
\includegraphics[width=\linewidth]{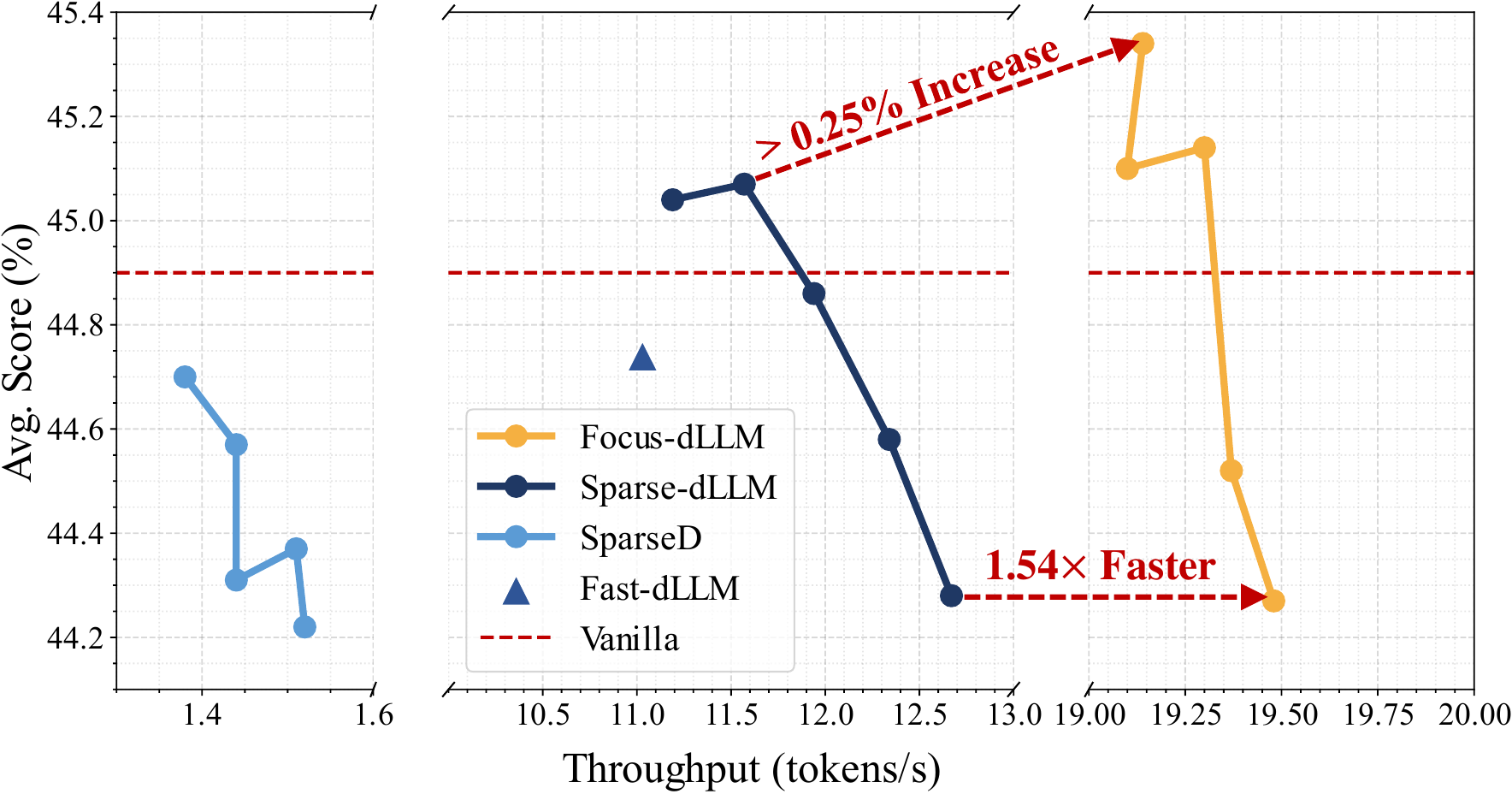} \caption{Accuracy \emph{vs.}\ throughput for UltraLLaDA~\cite{he2025ultralladascalingcontextlength} on LongBench~\cite{bai2024longbench} with $16K$. }
    \label{fig:pareto}
\end{figure}
Figure~\ref{fig:pareto} compares decoding throughput and LongBench~\cite{bai2024longbench} accuracy across different methods and configurations, with throughput measured at a $16K$ context length. Focus-dLLM consistently forms a stronger Pareto frontier than prior approaches, achieving higher throughput with comparable or better accuracy. Additional experimental details and configurations are provided in the Appendix~\ref{sec:details_pareto}.

\section{Ablation Study}
\begin{table}[!ht]
\centering
\caption{Ablation study of Focus-dLLM on UltraLLaDA~\cite{he2025ultralladascalingcontextlength}. PCGI denotes Past Confidence-Guided Indicator, and SA Sparse Attn represents sink-aware sparse attention.}
\label{tab:ablation_ultrallda}
\small
\setlength{\tabcolsep}{4pt}
\renewcommand{\arraystretch}{1.15}

\definecolor{hlgreen}{RGB}{223,241,222}
\definecolor{gain}{RGB}{0,150,0}
\definecolor{drop}{RGB}{150,0,0} 

\newcommand{\deltafmt}[2]{\hspace{2pt}{\scriptsize\textcolor{#1}{#2}}}
 \resizebox{0.76\linewidth}{!}{
\begin{tabu}{lcc}
\toprule
Method & Avg. Score & Throughput \\
\midrule

Fast-dLLM
& 44.74\deltafmt{gain}{\phantom{+0.00}}
& 11.03\deltafmt{gain}{\phantom{+0.00}} \\

\midrule
+ PCGI
& 44.23\deltafmt{drop}{-0.51}
& 11.37\deltafmt{gain}{+0.34} \\

+ SA Sparse Attn
& 44.84\deltafmt{gain}{+0.10}
& 17.68\deltafmt{gain}{+6.65} \\

\midrule
\rowcolor{mycolor!30}
Focus-dLLM
& \textbf{45.14}\deltafmt{gain}{+0.40}
& \textbf{17.71}\deltafmt{gain}{+6.68} \\
\bottomrule
\end{tabu}}
\end{table}
\noindent\textbf{Effectiveness of each component.}
We evaluate the impact of the proposed components on LongBench average score and $16K$ context decoding throughput. Table~\ref{tab:ablation_ultrallda} presents the ablation results building on the Fast-dLLM~\cite{wu2025fastdllmtrainingfreeaccelerationdiffusion} baseline. PCGI filters active queries via our past confidence-guided indicator while attending to the full context KV. SA Sparse Attn prunes context, while passing the entire block as active tokens and identifies redundancy for pruning using these tokens. Applying PCGI alone slightly degrades accuracy, while SA Sparse Attn improves accuracy by filtering irrelevant tokens in long contexts and significantly increasing throughput. Combining both components, Focus-dLLM achieves further accuracy gains and the highest throughput, demonstrating that accurate query selection enables more precise and effective sparse attention.

\begin{table}[!ht]
\centering
\caption{Effect of attention sinks on LongBench accuracy for Dream-7B-Instruct~\cite{ye2025dream7bdiffusionlarge}.
Incorporating attention sinks consistently improves performance across tasks.}
\label{tab:sink_ablation}
\small
\setlength{\tabcolsep}{4pt}
\renewcommand{\arraystretch}{1.12}

\definecolor{gain}{RGB}{0,150,0}

\begin{tabularx}{0.8\columnwidth}{l >{\raggedleft\arraybackslash}X >{\raggedleft\arraybackslash}X}
\toprule
Subset
& w/o Attn Sinks 
& w/ Attn Sinks \\
\midrule
hotpotqa  
& 37.17 & 38.96{\scriptsize\textcolor{gain}{+1.79}} \\
2wikimqa  
& 37.68 & 38.56{\scriptsize\textcolor{gain}{+0.88}} \\
trec      
& 69.50 & 70.00{\scriptsize\textcolor{gain}{+0.50}} \\
\midrule
Avg. Score & 41.47 & \textbf{42.82}{\scriptsize\textcolor{gain}{+1.35}} \\
\bottomrule
\end{tabularx}
\end{table}
Table~\ref{tab:sink_ablation} evaluates the effectiveness of attention sinks on
Dream-7B-Instruct~\cite{ye2025dream7bdiffusionlarge}. Incorporating attention
sinks leads to a clear improvement on LongBench~\cite{bai2024longbench}. These results suggest that effectively retaining attention sinks contributes to the preservation of key contextual information.

\begin{figure}[!ht]    
\includegraphics[width=\linewidth]{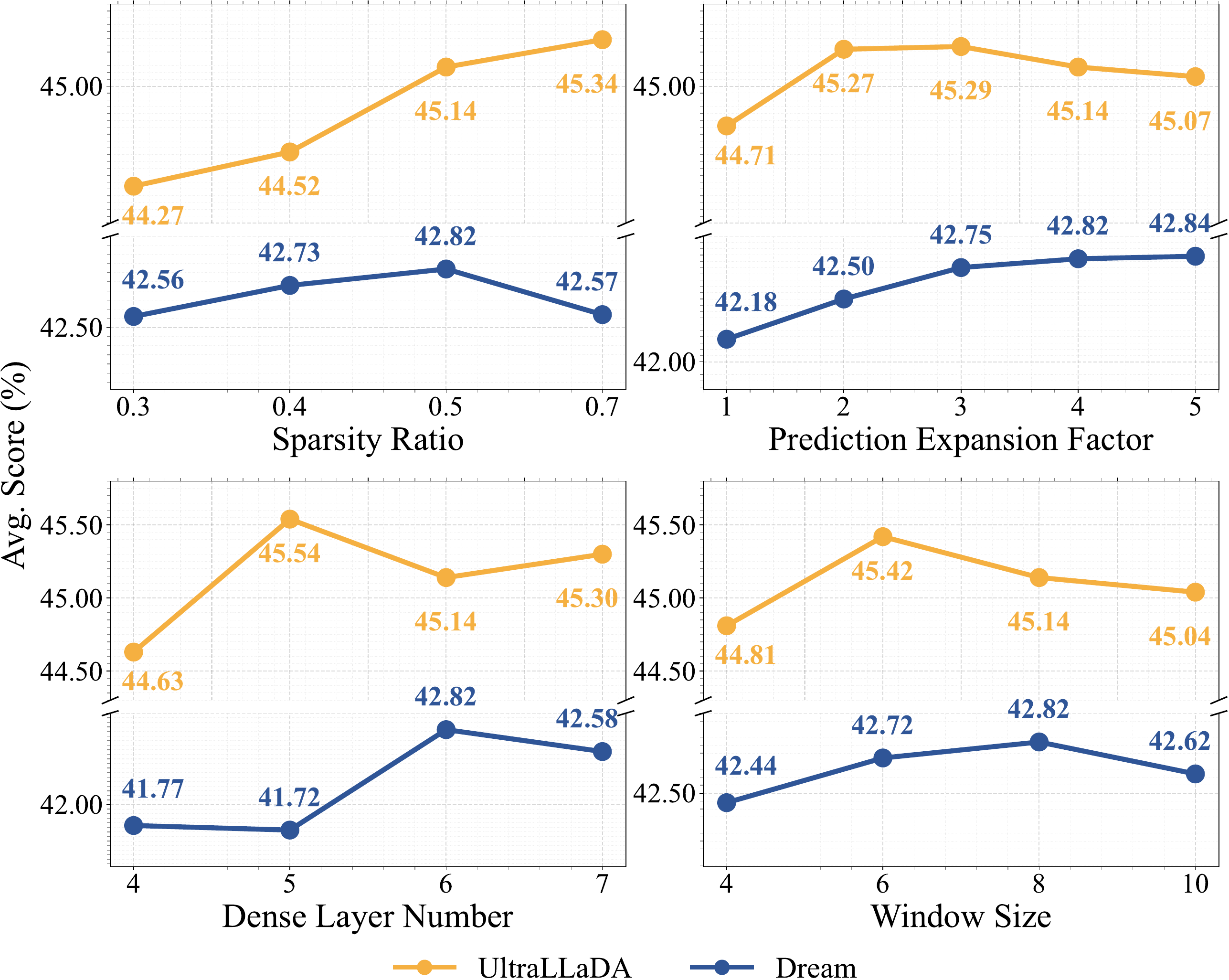} 
    \caption{Ablations on hyperparameters of Focus-dLLM on LongBench~\cite{bai2024longbench}.}
\label{fig:ablations}
\end{figure}

\noindent\textbf{Ablations on hyperparameters.}
Figure~\ref{fig:ablations} analyzes the impact of key hyperparameters in
Focus-dLLM. Increasing the sparsity ratio $\alpha$ generally improves accuracy,
indicating that retaining more relevant context benefits long-context reasoning,
while the drop observed for Dream~\cite{ye2025dream7bdiffusionlarge} at $\alpha{=}0.7$ suggests that excessive retention may introduce irrelevant context and dilute useful signals. For the prediction expansion factor $\rho$, small values (\emph{e.g.}, 1) lead to poor accuracy due to insufficient recall of future decoded token positions, whereas larger values provide more reliable coverage and steadily improve performance. Varying the number of dense layers $l_{\text{dense}}$ results in non-monotonic behavior, implying that attention sinks are not fully stabilized in shallow layers and that sparsification sensitivity differs across depths. A similar trend is observed for the window size $w$: overly small windows miss necessary local context, moderate windows improve accuracy, and excessively large windows degrade performance by introducing unrelated tokens.

\section{Conclusion}
We analyzed diffusion inference dynamics and introduced Focus-dLLM, a training-free framework for accelerating long-context dLLM inference. By leveraging a \textit{past confidence-guided indicator} for query prediction and a \textit{sink-aware pruning strategy} to retain critical history, our method effectively eliminates redundant computation. Experiments demonstrate that Focus-dLLM achieves over $29\times$ speedup at $32K$ context length while maintaining superior performance compared to state-of-the-art baselines.

\section*{Limitations}
While Focus-dLLM demonstrates high efficiency in text tasks, its extension to multimodal reasoning remains a direction for future exploration. Additionally, our current hyperparameters are manually configured, which may not achieve optimal performance across all specialized domains. Developing a fully adaptive mechanism for dynamic parameter adjustment represents a promising avenue to further enhance the framework's versatility and robustness.

\bibliography{custom}

\newpage
\begin{center}
    \Large{\textbf{Appendix}}
\end{center}
\appendix
\section{Implementation Details}
\label{sec:impl_details}
\subsection{Details of Focus-dLLM}
\label{sec:appendix_focus_details}

\begin{algorithm*}[!t]
\caption{Focus-dLLM Inference Procedure}
\label{alg:focus_dllm}
\begin{algorithmic}[1]
\Require Prompt $\mathbf{p}$, Mask token [MASK], Max steps $T$, Sparse ratio $\alpha$, Expansion factor $\rho$, Transfer Scheduler $\mathcal{S}$, Dense layers $l_{\text{dense}}$, Window size $w$.
\Ensure Generated sequence $\mathbf{x}^{(T)}$

\State Initialize $\mathbf{x}^{(0)} \leftarrow [\mathbf{p}, \text{[MASK]}_{1}, \dots, \text{[MASK]}_{N}]$
\State Initialize $\mathbf{K}, \mathbf{V}$ cache as empty; Confidence scores $\mathbf{c}^{(0)} \leftarrow \mathbf{0}$

\For{$t = 0$ to $T-1$}
    \State $\triangleright$ Determine dynamic prediction token counts
    \State $n^{(t)} \leftarrow$ Number of tokens to unmask at step $t$ given $\mathcal{S}$
    \State $k \leftarrow \lfloor \rho \cdot n^{(t)} \rceil$ \hfill $\triangleright$ Calculate candidate count

    \If{\text{IsBlockEntry}(t)}
        \State $\mathcal{I}_{\text{active}} \leftarrow \{1, \dots, L\}$ \hfill $\triangleright$ Full refresh at block entry
        \State $use\_sparse \leftarrow \text{False}$
    \Else
        \State $\mathcal{I}_{\text{focus}} \leftarrow$ Select top-$k$ indices based on $\mathbf{c}^{(t)}$ \hfill $\triangleright$ Candidate set
        \State $\mathcal{I}_{\text{active}} \leftarrow \bigcup_{i \in \mathcal{I}_{\text{focus}}} \{i - \lfloor w/2 \rfloor, \dots, i + \lfloor w/2 \rfloor\}$ \hfill $\triangleright$ Window expansion
        \State $use\_sparse \leftarrow \text{True}$
    \EndIf
    
    \State $\triangleright$ Layer-wise Forward Pass
    \For{layer $l = 1$ to $L_{layers}$}
        \If{$use\_sparse$ \textbf{and} $l > l_{\text{dense}}$}
            \State $\triangleright$ Sparse Attention Mechanism
            \State $\mathcal{I}_{\text{sink}} \leftarrow \text{IdentifySinks}(\text{Layer } l_{\text{dense}})$
            \State Compute Block Relevance $R_b$ using $\mathbf{Q}_{\mathcal{I}_{\text{focus}}}$ and $\bar{\mathbf{K}}_b$
            \State Determine selection size $C = \lfloor \alpha \cdot N_{\text{total\_blocks}} \rfloor$
            \State Select relevant prompt blocks $\mathcal{B}_{\text{relevant}} \leftarrow \text{Top-C}(R_b)$
            
            \State $\mathcal{I}_{p} \leftarrow \mathcal{I}_{\text{sink}} \cup \bigcup_{b \in \mathcal{B}_{\text{relevant}}} \{\, i \mid i \in \text{Block}_b \,\}$
            \State $\mathbf{K}_{\text{attn}} \leftarrow \text{Concat}(\mathbf{K}_{\mathcal{I}_p}, \mathbf{K}_{\text{resp}})$
            \State $\mathbf{V}_{\text{attn}} \leftarrow \text{Concat}(\mathbf{V}_{\mathcal{I}_p}, \mathbf{V}_{\text{resp}})$
            
            \State $\mathbf{H}_l \leftarrow \text{Softmax}\left(\frac{\mathbf{Q}_{\mathcal{I}_{\text{active}}} \mathbf{K}_{\text{attn}}^\top}{\sqrt{d}}\right) \mathbf{V}_{\text{attn}}$
        \Else
            \State $\triangleright$ Full Attention \& Cache Update
            \State $\mathbf{H}_l \leftarrow \text{FullAttn}(\mathbf{Q}_{\mathcal{I}_{\text{active}}}, \mathbf{K}, \mathbf{V})$
            \State Update KV Cache for indices in $\mathcal{I}_{\text{active}}$
        \EndIf
    \EndFor
    
    \State $\triangleright$ Denoising and State Update
    \State Update $\mathbf{x}^{(t)}$ to $\mathbf{x}^{(t+1)}$ and compute new confidence $\mathbf{c}^{(t+1)}$
\EndFor
\State \Return $\mathbf{x}^{(T)}$
\end{algorithmic}
\end{algorithm*}

\begin{table*}[!ht]
\centering
\caption{Detailed information of the datasets in the LongBench benchmark.}
\label{tab:longbench_datasets}
\resizebox{0.9\linewidth}{!}{\begin{tabu}{l l r r r r r}
\toprule
Label & Eval. Metric & Avg. Len. & Gen. Len. & Steps & Language & Sample Num. \\
\midrule
Qasper     & F1        & 3,619  & 32  & 32  & EN              & 200 \\
MultiFieldQA-en & F1        & 4,559  & 64  & 64  & EN              & 150 \\
HotpotQA   & F1        & 9,151  & 32  & 32  & EN              & 200 \\
2WikiMQA   & F1        & 4,887  & 32  & 32  & EN              & 200 \\
Musique    & F1        & 11,214 & 32  & 32  & EN              & 200 \\
GovReport  & Rouge-L   & 8,734  & 512 & 512 & EN              & 200 \\
QMSum      & Rouge-L   & 10,614 & 512 & 512 & EN              & 200 \\
MultiNews  & Rouge-L   & 2,113  & 512 & 512 & EN              & 200 \\
TREC       & Accuracy  & 5,177  & 64  & 64  & EN              & 200 \\
TriviaQA   & F1        & 8,209  & 32  & 32  & EN              & 200 \\
SAMSum     & Rouge-L   & 6,258  & 128 & 128 & EN              & 200 \\
Lsht       & Accuracy  & 22,333 & 64  & 64  & ZN              & 200 \\
PassageRetrieval        & Accuracy  & 9,289  & 32  & 32  & EN              & 200 \\
Lcc        & Edit Sim  & 1,235  & 64  & 64  & Python/C\#/Java & 500 \\
RepoBench-P       & Edit Sim  & 4,206  & 64  & 64  & Python/Java     & 500 \\
\bottomrule
\end{tabu}}
\end{table*}

This section provides additional details regarding the implementation of our Focus-dLLM framework.
To maximize computational efficiency, our framework leverages specialized GPU kernels for attention computations. The core sink-aware sparse attention operator, which handles dynamic context pruning, is implemented using Triton~\cite{tillet2019triton}. This allows for fine-grained control and optimization of memory access patterns for sparse matrix operations. For dense attention computations—which occur in the initial dense layers ($l \le l_{\text{dense}}$) and during full-cache refreshes at block entries—we utilize the highly optimized FlashAttention~\cite{shah2024flashattention} kernel to accelerate inference.

Empirical analysis revealed that the final layers of Dream-7B-Instruct~\cite{ye2025dream7bdiffusionlarge} exhibit high sensitivity to attention sparsification. To mitigate potential performance degradation, we designate the final four transformer layers of this model as dense, ensuring they always perform full attention. This hybrid strategy preserves the integrity of critical generation stages, achieving a superior trade-off between accuracy and efficiency.

Focus-dLLM strictly adhere to the original decoding strategies of both UltraLLaDA~\cite{he2025ultralladascalingcontextlength} and Dream-7B-Instruct~\cite{ye2025dream7bdiffusionlarge}. The generation process follows the semi-autoregressive remasking paradigm, where a transfer scheduler dictates which tokens are unmasked at each step based on their confidence scores, consistent with the methods described in ~\cite{he2025ultralladascalingcontextlength} and ~\cite{ye2025dream7bdiffusionlarge}. The complete inference procedure of Focus-dLLM is detailed in Algorithm~\ref{alg:focus_dllm}.

\subsection{Baselines}

In our experiments, we compare Focus-dLLM against the vanilla inference of representative diffusion models and several state-of-the-art acceleration frameworks.

\noindent{\textbf{Vanilla dLLMs.}
We use the standard inference implementations of UltraLLaDA \citep{he2025ultralladascalingcontextlength} and Dream-7B-Instruct \citep{ye2025dream7bdiffusionlarge} as our primary baselines. UltraLLaDA is developed by fine-tuning LLaDA~\cite{nie2025llada}  for long-context capabilities, while Dream is adapted from a pre-trained autoregressive model. These representative diffusion LLMs perform a full attention computation over the entire sequence at each denoising step, without any caching or sparsification mechanisms.

\noindent{\textbf{Fast-dLLM.}
As a strong baseline for approximate KV cache methods, Fast-dLLM \citep{wu2025fastdllmtrainingfreeaccelerationdiffusion} introduces a block-wise approximate KV cache tailored for the bidirectional attention in dLLMs. It reuses cached activations from previously decoded blocks to reduce redundant computation.

\noindent{\textbf{Sparse-dLLM.}
This method \citep{song2025sparsedllmacceleratingdiffusionllms} accelerates dLLM inference by integrating dynamic cache eviction with sparse attention principles. It leverages the temporal stability of token saliency to identify and retain critical KV entries while dynamically evicting unimportant entries from both the prefix and suffix contexts. 

\noindent{\textbf{SparseD.}
As a pure sparse attention baseline, SparseD \citep{wang2025sparsed} is tailored for the unique attention patterns in dLLMs. Its core strategy involves pre-computing head-specific sparse patterns once and reusing them across subsequent denoising steps. To preserve generation quality, it applies full attention during the critical early steps before switching to the pre-computed sparse patterns for the remainder of the inference process.

To ensure a fair comparison, all methods uniformly employ the semi-autoregressive remasking strategy, with the block length set to 32 across all experiments. 

\subsection{Generation Settings}
Table~\ref{tab:longbench_datasets} presents the detailed configurations for each task in the LongBench benchmark. To align with the evaluation settings of UltraLLaDA \citep{he2025ultralladascalingcontextlength}, we process all input contexts by truncating them to a maximum length of $16K$ tokens using the "drop-middle" strategy. For each specific task, the generation length (Gen. Len.) and the generation steps (Steps) are configured as specified in the table.

\section{Details of Accuracy \emph{vs.} Efficiency Experiments}
\label{sec:details_pareto}

\begin{table*}[!ht]
\centering
\scriptsize
\setlength{\tabcolsep}{2pt}
\renewcommand{\arraystretch}{1.15}
\caption{Detailed performance and throughput comparison on LongBench~\cite{bai2024longbench} for UltraLLaDA~\cite{he2025ultralladascalingcontextlength}. We report results for baselines and various configurations of our method, Focus-dLLM.}
\label{tab:focus_dllm_pareto}
\resizebox{\linewidth}{!}{
\begin{tabular}{l*{15}{c}}
\toprule
\multirow{5}{*}{Method}
& \multicolumn{2}{c}{Single-Doc. QA}
& \multicolumn{3}{c}{Multi-Doc. QA}
& \multicolumn{2}{c}{Summarization}
& \multicolumn{2}{c}{Few-shot Learning}
& \multicolumn{2}{c}{Synthetic}
& \multicolumn{2}{c}{Code}
& \multirow{5}{*}{Ave.\ Score}
& \multirow{5}{*}{\begin{tabular}{c} Throughput(16K) \end{tabular}}
\\
\cmidrule(lr){2-3}
\cmidrule(lr){4-6}
\cmidrule(lr){7-8}
\cmidrule(lr){9-10}
\cmidrule(lr){11-12}
\cmidrule(lr){13-14}
& \rotatebox{60}{Qasper}
& \rotatebox{60}{MF-en}
& \rotatebox{60}{HotpotQA}
& \rotatebox{60}{2WikiMQA}
& \rotatebox{60}{Musique}
& \rotatebox{60}{GovReport}
& \rotatebox{60}{QMSum}
& \rotatebox{60}{TREC}
& \rotatebox{60}{TriviaQA}
& \rotatebox{60}{LSHT}
& \rotatebox{60}{PRe}
& \rotatebox{60}{Lcc}
& \rotatebox{60}{RB-P}
& \\
\midrule

Vanilla
& 19.14 & 25.87 & 16.27 & 18.00 & 12.08
& 32.83 & 22.48 & 80.00 & 91.58 & 41.00
& 96.75 & 68.23 & 59.50
& 44.90 & 1.06 \\ 

Fast-dLLM
& 18.34 & 29.90 & 17.03 & 17.11 & 13.36
& 30.05 & 22.89 & 79.50 & 91.03 & 42.00
& 94.75 & 67.50 & 58.10
& 44.74 & 11.03 \\

\midrule

Sparse-dLLM (r=0.3)
& 17.10 & 25.82 & 20.82 & 18.63 & 15.35
& 27.95 & 22.52 & 71.00 & 91.93 & 41.50
& 98.71 & 67.07 & 57.18
& 44.28 & 12.67\\

Sparse-dLLM (r=0.4)
& 17.99 & 27.67 & 18.90 & 18.55 & 13.11
& 28.77 & 23.01 & 74.50 & 91.43 & 42.00
& 98.17 & 67.80 & 57.68
& 44.58 & 12.34\\

Sparse-dLLM (r=0.5)
& 18.04 & 27.26 & 20.59 & 17.88 & 13.67
& 29.95 & 23.57 & 76.50 & 91.93 & 41.50
& 97.12 & 67.50 & 57.72
& 44.86 & 11.94\\

Sparse-dLLM (r=0.6)
& 19.06 & 26.94 & 20.80 & 18.30 & 14.31
& 30.16 & 23.68 & 77.00 & 91.43 & 42.50
& 96.25 & 67.99 & 57.44
& 45.07 & 11.57\\

Sparse-dLLM (r=0.7)
& 19.03 & 27.35 & 21.64 & 18.04 & 13.24
& 30.63 & 23.38 & 78.50 & 91.43 & 41.50
& 95.42 & 67.94 & 57.40
& 45.04 & 11.19\\

\midrule

SparseD (skip=0.2,r=0.3)
& 19.09 & 25.87 & 15.45 & 18.04 & 11.92
& 32.64 & 22.50 & 79.50 & 90.70 & 41.50
& 96.79 & 68.10 & 59.02
& 44.70 & 1.38\\

SparseD (skip=0.2,r=0.2)
& 18.85 & 25.70 & 16.10 & 17.14 & 11.64
& 32.29 & 22.52 & 79.50 & 90.70 & 41.50
& 96.67 & 68.02 & 58.77
& 44.57 & 1.44\\

SparseD (skip=0.2,r=0.1)
& 18.06 & 25.76 & 15.40 & 16.64 & 11.47
& 31.44 & 22.60 & 79.50 & 91.05 & 41.50
& 96.84 & 67.98 & 58.59
& 44.37 & 1.51\\

SparseD (skip=0.1,r=0.3)
& 18.89 & 24.72 & 14.45 & 14.65 & 10.93
& 32.55 & 22.93 & 79.50 & 91.70 & 41.50
& 97.12 & 67.86 & 59.18
& 44.31 & 1.44\\

SparseD (skip=0.1,r=0.2)
& 19.34 & 24.21 & 14.09 & 13.84 & 10.80
& 31.61 & 23.12 & 79.50 & 91.53 & 42.00
& 97.88 & 67.72 & 59.16
& 44.22 & 1.52\\

\midrule

\rowcolor{mycolor!30}
Focus-dLLM ($\alpha$=0.3)
& 16.63 & 29.57 & 23.09 & 21.14 & 18.59
& 26.09 & 21.05 & 69.50 & 91.28 & 38.00
& 97.27 & 66.19 & 57.07
& 44.27 & 19.48\\

\rowcolor{mycolor!30}
Focus-dLLM ($\alpha$=0.4)
& 16.60 & 27.83 & 23.81 & 21.34 & 18.52
& 26.85 & 21.03 & 72.00 & 90.78 & 40.00
& 97.23 & 66.55 & 56.18
& 44.52 & 19.37\\

\rowcolor{mycolor!30}
Focus-dLLM ($\alpha$=0.5)
& 17.02 & 29.11 & 22.47 & 21.49 & 20.20
& 26.75 & 21.45 & 77.00 & 90.78 & 41.00
& 95.73 & 66.72 & 57.12
& 45.14 & 19.30\\

\rowcolor{mycolor!30}
Focus-dLLM ($\alpha$=0.6)
& 16.91 & 28.20 & 22.75 & 23.21 & 18.94
& 26.39 & 21.86 & 76.50 & 90.78 & 41.50
& 95.84 & 66.67 & 56.75
& 45.10 & 19.10\\

\rowcolor{mycolor!30}
Focus-dLLM ($\alpha$=0.7)
& 17.71 & 29.36 & 23.61 & 22.52 & 19.12
& 26.97 & 21.58 & 76.50 & 90.78 & 41.00
& 96.67 & 66.85 & 56.72
& 45.34 & 19.14\\

\bottomrule
\end{tabular}}
\end{table*}

This section provides the detailed results underpinning our accuracy vs. efficiency analysis. Table~\ref{tab:focus_dllm_pareto} presents a comprehensive performance comparison on the LongBench~\cite{bai2024longbench} for UltraLLaDA~\cite{he2025ultralladascalingcontextlength}, including various configurations for both baseline methods and our own. For Sparse-dLLM \citep{song2025sparsedllmacceleratingdiffusionllms}, we vary the retention ratio r, which determines the percentage of KV cache entries preserved. For SparseD \citep{wang2025sparsed}, configurations differ in the skip ratio (the initial portion of steps using full attention) and the selection ratio r. The configurations for our method, Focus-dLLM, correspond to different settings of the sparsity ratio $\alpha$ , which controls the amount of prompt context retained for attention computation, while all other hyperparameters remain consistent with the setup described in the main text(section  ~\ref{sec:exp_settings}). As the results consistently demonstrate, Focus-dLLM establishes a better accuracy-efficiency trade-off, achieving superior overall performance compared to prior acceleration methods.

\section{Attention Patterns of dLLM}

\begin{figure*}[t]
\centering
\includegraphics[width=0.8\linewidth]{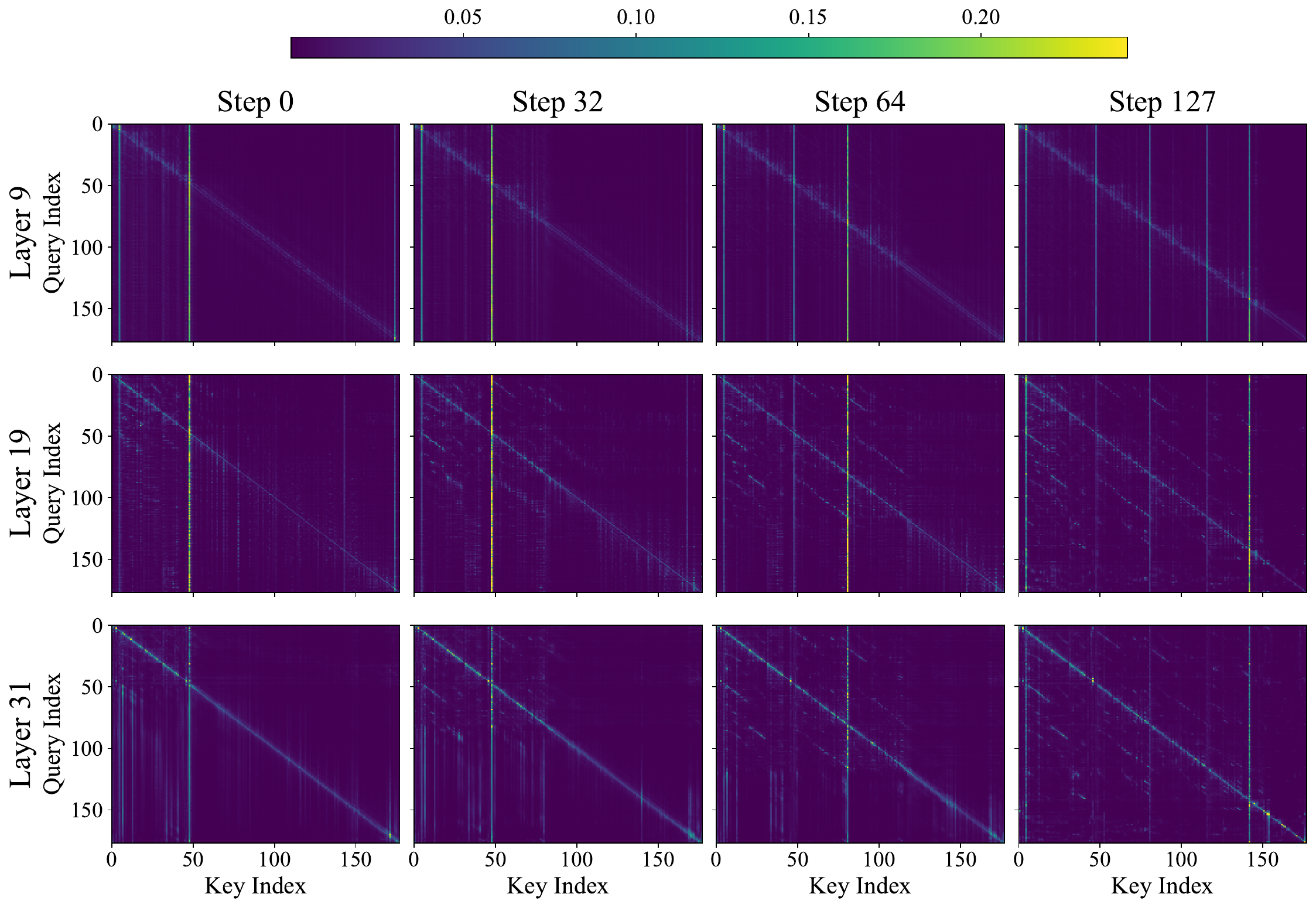} 
    \caption{Attention patterns in LLaDA-8B-Instruct~\cite{nie2025llada} across various layers and denoising steps. The heatmaps demonstrate the emergence of attention sinks (vertical bands) and their strong positional consistency across different layers within the same step.}
\label{fig:attention_grid_full}
\end{figure*}

To supplement our analysis in Section~\ref{sec:dsap}, Figure~\ref{fig:attention_grid_full} presents a broader visualization of attention patterns from LLaDA-8B-Instruct~\cite{nie2025llada} across various layers and denoising steps. The heatmaps clearly illustrate the principles of locality (strong diagonals) and the formation of attention sinks (bright vertical bands). Most importantly, the figure provides strong visual evidence for the cross-layer consistency of these sinks. The locations of the prominent vertical bands are remarkably stable across different layers (\emph{e.g.}, Layer 9, 19, and 31) at any given denoising step. This observed stability is the primary motivation behind our method, as it validates our strategy of identifying sink locations at an intermediate depth and reusing them for deeper layers to eliminate redundant computation.


\end{document}